\documentclass[letterpaper, 10 pt, conference]{IEEEtran}  

\usepackage{amsmath,amssymb,amsfonts}

\usepackage{algorithm}
\usepackage{algorithmic}

\usepackage{graphicx}
\usepackage{textcomp}
\usepackage{xcolor}
\usepackage{amsmath}
\usepackage{float}
\usepackage{amsfonts}
\usepackage{bbm}
\usepackage{bbm}
\usepackage{graphicx}
\usepackage{varwidth}
\usepackage{caption}
\let\labelindent\relax
\usepackage{enumitem}
\usepackage{dsfont}
\usepackage{tabularx}
\usepackage{booktabs}
\usepackage{subfigure}

\usepackage{pgfplots}
\usepgfplotslibrary{colorbrewer}
\pgfplotsset{compat = 1.14, cycle list/Set1-8} 
\usetikzlibrary{pgfplots.statistics, pgfplots.colorbrewer,pgfplots.dateplot,
        shadows,
        matrix} 
\usepackage{pgfplotstable}
\usepackage{filecontents}

\usepackage{graphicx}
\pgfplotsset{compat=1.14}

\title{\LARGE \bf
Mechanisms for Integrated Feature Normalization and Remaining Useful Life Estimation Using LSTMs Applied to Hard-Disks
}

\author{ \parbox{6 in}{\centering Sanchita Basak, Saptarshi Sengupta, Abhishek Dubey\\
        Department of Electrical Engineering and Computer Science, Vanderbilt University\\
        Nashville, TN, USA\\
        {\{sanchita.basak, saptarshi.sengupta, abhishek.dubey\}@vanderbilt.edu}}
}

\begin{document}

\maketitle
\thispagestyle{empty}
\pagestyle{empty}


 


\begin{abstract}






In this paper we focus on application of data-driven methods for remaining useful life estimation in components where past failure data is not uniform across devices, i.e. there is a high variance in the minimum and maximum value of the key parameters. The system under study is the hard disks used in computing cluster. The data used for analysis is provided by Backblaze as discussed later. 
 In the article, we discuss the architecture of of the long short term neural network used and describe the mechanisms to choose the various hyper-parameters. Further, we describe the challenges faced in extracting effective training sets from highly unorganized and class-imbalanced big data and establish methods for online predictions with extensive data pre-processing, feature extraction and validation through online simulation sets with unknown remaining useful lives of the hard disks. Our algorithm performs especially well in predicting RUL near the critical zone of a device approaching failure. With the proposed approach we are able to predict whether a disk is going to fail in next ten days with an average precision of 0.8435. We also show that the architecture trained on a particular model is generalizable and transferable as it can be used to predict RUL for devices in other models from same manufacturer. 


\end{abstract}

\maketitle
\pagenumbering{arabic}
\thispagestyle{empty}
\pagestyle{plain}



\section{Introduction}

In the past we have developed approaches for both model-driven as well as data driven prognostics as shown by our work on the predictive health maintenance of lithium ion batteries employed in small satellite missions discussed in \cite{Sun2018ADD}. However, most of those past work were contingent about uniform failure metrics as observed across the devices. For example, in \cite{Sun2018ADD} we discovered that the battery failure semantics remained uniform across different devices. However, in some cases, the same metric when measured across  different devices show variance in the minimum and maximum value observed. For example, we will show later in the paper that the hard disks from the same manufacturer show large variance when we look at the observed SMART parameters, especially near the point of failure. However, the interesting point to note is that always the same SMART parameters are critical markers of failures.

These observations lead to an interesting challenge in training the remaining useful life (RUL) predictors while normalizing the data, when the range of values the features can take, vary vastly among the devices.  To showcase this approach, the Backblaze dataset \cite{WinNT} is considered. There have been several studies directed at predicting RUL of hard disks using model-driven approaches \cite{Schroeder:2007:DFR:1267903.1267904}. However, they cannot always capture the device dynamics well \cite{8023108} using standard distributions specially when it is important to consider the trend of variance in the minimum and maximum range of the feature values across the devices. This is required in order to have an effective normalization strategy for prediction. 

\textbf{Contributions:} To account for these challenges, 
this paper develops an online prediction model using Deep Long Short Term Memory (LSTM) \cite{doi:10.1162/neco.1997.9.8.1735} networks. It is motivated by the need to provide accurate predictive analytics despite the unusually non-linear pattern dynamics of the Backblaze hard drive dataset \cite{WinNT}. The goal, in summary, is to predict the remaining useful life of the drives under test in the Backblaze data. The results presented in the paper show an average precision of 0.8435, recall of 0.72, and F1 score of 0.77 when predicting whether a hard disk is going to fail in the next ten days. The  contributions discussed in the paper are:
\begin{itemize}[leftmargin=*]
\item{\textbf{Learning failure patterns using device-specific normalization:}} It should be noted that even if the devices worked with are from the same manufacturer, they do not necessarily fail under similar conditions i.e. with similar failure-specific feature values. To make things worse, at some of the feature values deemed healthy for one device, another device may fail. It should be noted that the data has some failure states which have similar feature sets to those of active states \cite{8260700}. This renders the possibility of finding out one specific set of feature values corresponding to global failure slim. This in turn makes it infeasible for traditional Machine Learning (ML) paradigms to learn the highly non-linear causal embeddings latent in each instance of progression towards a fault. In the proposed approach we have taken care of extracting the training set from such highly unorganized feature sets with major class imbalances and established custom, device-specific normalization techniques in the training and testing stages to overcome this issue. 
\item {\textbf{Generalization of prediction model within the class of devices:}} Most of the existing work in this domain uses cross validation. However, we divide the data into \textit{training}, \textit{validation}, and a \textit{simulation} set (for testing) such that the training and validation data are from the same distribution where the RUL of a device is known, but the simulation data is the online data coming in real time for devices with unknown RUL. This enables extension of prediction capacity to simulation data being fed in an online manner whose failure logs have not been used in the training process - the caveat is to preprocess the data for such devices in such a way so that they have similar underlying pattern mappings to that of the training set.  Further, we show that our trained models work well even for other hard disk models from the same manufacturer. 

\end{itemize}

The rest of the paper is organized as follows:  \text{Section~\ref{sec:problemformulation}} outlines the problem. \text{Section~\ref{sec:relatedwork}} discusses the related work. \text{Section~\ref{sec:approach}} describes our approach. \text{Section~\ref{sec:lstms}} describes the Long Short Term Memory Architecture used for the prediction framework. \text{Section~\ref{sec:results}} reports the results followed by an involved analysis of the outcomes, \text{Section~\ref{sec:comparative}} discusses the comparative analyses with the existing research while \text{Section~\ref{sec:conclusions}} provides concluding remarks and future directions.





\section{Problem Description}
\label{sec:problemformulation}

\begin{figure}[h]
\includegraphics[width=\columnwidth, height=5.6cm]{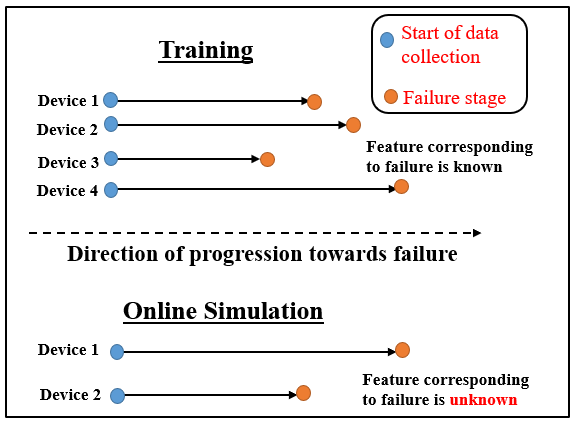}
\centering
\caption{General scenario on training and online simulation data}
\label{fig:general}
\end{figure}

The generalized problem that we are trying to solve is as described in \text{Figure~\ref{fig:general}}. We have training data for several devices from the start of data collection upto failure stage. Different devices fail at different time instants with different feature values. Some of the feature values deemed healthy  for  one  device, could cause failure in another device. For the training set, the features corresponding to failure is known. But as there is no universal feature set causing failures across all devices we go for device specific normalizations with respect to the failure causing features for preprocessing of training data before feeding them into the LSTM networks. 

For the online simulation data, features corresponding to failure are not known apriori. Simulation data comes in real time for devices that are active now but going to fail sometime in future. We propose mechanisms for integrated feature normalization for online simulation data to estimate failure causing features with respect to which the data needs to be normalized so that they have similar underlying pattern mappings to that of training set. Only then the trained network can be applied on simulation data to accurately predict RUL.

\subsection{\textbf{Dataset}}
\label{sec:datades}
We are working with the Backblaze hard drive dataset and providing an approach to the solution to the problem described above. The dataset contains 
 the snapshot of 30 different S.M.A.R.T. (Self-Monitoring, Analysis and Reporting Technology) indices including both raw and normalized values for each operational  hard  drive  model  from  various  companies reported once every twenty-four hours has been considered for this work.

We captured the data sequences from January 2017 through December 2017 containing information about 91,243 devices manufactured by various companies out of which we chose to work with the device model ST4000DM000 from Seagate due to the following reasons: 

\begin{itemize}[leftmargin=*]
\item Failure statistics from 2017 suggest Seagate devices failed the most 
\item Out of all device models ST4000DM000 from Seagate contributed to most of the failures.
\end{itemize}



\begin{table}[h!]
\caption{Summary of SMART features Used}
\centering
 \begin{tabular}{ | m{0.7cm} | m{0.8cm} | m{5.5 em}| m{12.3 em} | } 
 \hline
 Feature no. & SMART ID & Attribute Name & Description \\  
 \hline\hline
\centering5 & \centering7 & \centering Seek error rate & Frequency of the errors during disk head positioning and rises with approaching failure. \\ 
 \hline
 \centering6 & \centering9 & \centering Power-on-hours count & Estimated remaining lifetime, based on the time a device was powered on. \\
 \hline
 \centering22 & \centering240  & \centering Head flying hours/transfer error-rate & Time spent during the positioning of the drive heads \\
 \hline
 \centering23 & \centering241 & \centering Total LBAs written & Related to the use and hence indicating the aging process of hard drives \\
 \hline
 \centering24 & \centering242 & \centering Total LBAs read & Related to the use and hence indicating the aging process of hard drives \\  
 \hline

 \end{tabular}
 \label{tab:selectedfeatures}
\end{table}

\begin{table}[h]
\caption{Summary of Symbols Used}
\centering
 \begin{tabular}{ | m{1.5cm} | m{20em}| } 
 \hline
 Symbols & Meaning  \\  
 \hline\hline

 ${f}$ & Selected features \\
 \hline
 ${A_i}$ & Training data matrix for each device ${i}$  \\  
 \hline
${T_0}$  & Day when the data collection starts  \\ 
 \hline
 ${T_f}$ & Day when the device fails\\
 \hline
 
 ${T_c}$ & The  current  time  when  the  device  is  active and when we want to test the device for its RUL\\
 \hline
 ${H_{sh\times f}}$  & Historical data ${H}$ of past two months from ${T_c}$  with ${sh}$ sample size and ${f}$ features \\
 \hline
 ${\phi}$ & Sorted ${H}$ in ascending order \\
 \hline
 
 ${Q_3}$ & Set of all elements contained between 50th and 75th percentile of data\\
 \hline
 ${\phi_{75}}$ & 75th percentile of Sorted ${H}$ \\
 \hline

 ${B_i}$ & Simulation data matrix for each device ${i}$  \\  
 \hline
 
 ${\Tilde{B_i}}$ & Normalized ${B_i}$ \\  
 \hline
 
 ${j}$ & Time index for training data varying from  ${T_f-150,.....,T_f}$\\
 \hline
 ${m}$ & Time instances for simulation data varying from  ${T_c-150,.....,T_c}$  \\  
 \hline
 ${F_n}$  & Fisher score of feature ${n}$  \\
 \hline
 ${\mu _n ^k}$  & Mean of the ${n_{th}}$ feature for class ${k}$  \\
 \hline
 ${\sigma _n ^k}$  & Variance of the ${n_{th}}$ feature for class ${k}$ \\
 \hline
 ${d_k}$  &  The  number  of  data  instances  in  each  class   \\
 \hline
 ${ts}$  &  Time steps to look back in the LSTM network which is considered to be 25  \\
 \hline
 
 \end{tabular}
 \label{tab:symbolsummary}
 
\end{table}







\subsection{Related Work}

\label{sec:relatedwork}

Classical approaches on error prediction of disk drives are generally model-driven. They focus on modelling the failure patterns with statistical distributions. B. Schroeder et al.\cite{Schroeder:2007:DFR:1267903.1267904},  provided a quantitative analysis of hard disk replacement rates and discussed the statistical properties of the distribution capturing time between replacements. On the other hand, Wang et al. \cite{8023108}, concludes that the time between failure cannot be captured through any of the standard distributions and commented on the difficulty of capturing fault trend using standard distributions.

The recent works on device health forecasting has adopted an array of approaches: in one study by Eker et al. \cite{Eker2014ASP} RUL prediction was carried out by directly comparing sensor similarity instead of using any health estimates. 
With the thrust moving towards data-driven approaches Gugulothu et al. \cite{Gugulothu2017PredictingRU} used an RNN model to generate embeddings which capture the summary trend of multivariate time series data followed by factoring the notion that embeddings for healthy and degraded devices tend to be different, into their forecasting scheme. Recent work on device health monitoring as evidenced in \cite{Malhotra2016LSTMbasedEF} reinforce the idea of using RNNs to capture intricate dependencies among sensor observations across time cycles of dynamic period range.  In \cite{Botezatu:2016:PDR:2939672.2939699}, the authors came up with disk replacement prediction algorithm with changepoint detection in time series Backblaze data and concluded some rules for directly identifying the state of a device: healthy or faulty. Aussel et al., \cite{8260700} used the same dataset to perform hard drive failure prediction with SVM, RF and GBT and discussed their performances based on precision and recall. Prediction of remaining useful lives  using quantum particle swarm optimization \cite{8301693} \cite{2018arXiv180702870S} of lithium-ion battery has been discussed in \cite{Yu2017} and a host of recent swarm intelligence algorithms \cite{Sengupta2018ParticleSO}  can be effectively applied in  prediction of RUL of various devices in conjuction with other ML approaches. We present a comparative analysis with some of these works in Section \ref{sec:comparative}.

\section{Our Approach}
\label{sec:approach}


In this work, we seek to provide a RUL prediction model by using a Deep Long Short Term Memory (LSTM) \cite{doi:10.1162/neco.1997.9.8.1735} network. We selected 5 features out of 24 as shown in \text{Table~\ref{tab:selectedfeatures}}, through feature selection methods discussed later in this section. If the data collection starts at day ${T_0}$ and a device can fail in any day ${T_f}$ after that, a device can have data for all such time indices ${j}$ where ${j}$  varies from  ${T_0}$ to ${T_f}$. All the time instances have been discretely sampled at an interval of one day. We can formulate a multivariate time series data ${A_i\subseteq \mathbb{R}^{j\times f}}$ (${f}$ being the selected features) for each device ${i}$. So ${A_i}$ should have past ${j}$ days of features starting from the day of failure ${T_f}$ until ${T_0}$. For normalization of feature matrices  we need to have similar length for each ${A_i}$ data matrix, such that the time index ${j}$ varies from ${T_f}$ upto a fixed length of a sequence back, which in our case is upto 150 days prior to failure as we are interested to find failures occurring within this time region. Now the training dataset can be represented as:  \{${A_i\subseteq \mathbb{R}^{j\times f}\mid  j={T_f}, {T_f}-1,....,{T_f}-150}$\}. The corresponding training labels attached to each row of ${A_i}$ is:  \{${T_f-j}$\}.

For online simulation data, we consider ${T_c}$ as the current time when the device is active and when we want to test the device for its Remaining Useful Life, such that ${T_0<T_c<T_f}$. Hence the problem boils down to predicting the RUL of a device ${i}$, i.e., the remaining time it is active from the time instant ${T_c}$ upto failure. Hence, a simulation dataset can be formulated such that 
\{${B_i\subseteq \mathbb{R}^{m\times f}\mid  m={T_c}, {T_c}-1,....,{T_c}-150}$\} where ${B_i}$ is the simulation data matrix for each device ${i}$ with data for varying time index ${m}$. \text{Table~\ref{tab:symbolsummary}} lists all the symbols used in this paper along with their description.


\subsection{\textbf{Feature selection}}

\begin{figure}[h]
\includegraphics[width=\columnwidth]{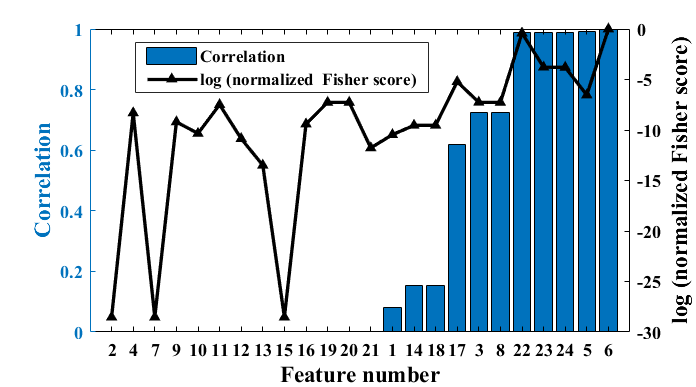}
\centering
\caption{Sorted feature list in ascending order of correlation and Fisher score value. Table \ref{tab:selectedfeatures} summarizes the five chosen features.}
\label{fig:correlation}
\end{figure}

All the twenty four SMART statistics reported by the Seagate model are not correlated with the progression towards failure. 
First, correlation coefficients between each individual feature and the RUL are calculated to observe how the feature trends change as the device progresses towards failure and the features with highest absolute values of correlation score are selected.
  \text{Figure~\ref{fig:correlation}} shows the correlation of each feature with failure. The features are sorted according to increasing order of correlation values. Five out of twenty four features have been chosen in this way with feature numbers shown in the plot.
 
Next we choose another supervised filter feature selection method, Fisher score \cite{SHEIKHPOUR2017141}, which focuses on features having better distinguishing capability in terms of greater variance of values among different classes and more similar feature values within a particular class. Let us consider, an input dataset ${x}$ having ${n}$ features with their labels ${y}$ comprising of ${L}$ different classes with class index ${k}$ such that ${k = 1,2,...L}$ and the number of data instances in each class is  ${d_k}$. We calculate the mean of all data samples of ${n_{th}}$ feature as ${\mu _n}$,  while ${\mu _n ^k}$ and ${\sigma _n ^k}$ are the mean and variance of the ${n_{th}}$ feature for class ${k}$. Hence the Fisher score of ${n_{th}}$ feature ${F_n}$ is described as:

\begin{equation}
F_n = \frac{\sum_{k=1}^{L}d_k(\mu _n ^k - \mu _n)^2}{\sum_{k=1}^{L}d_k(\sigma _n ^k)^2}
\end{equation}

The numerator ${\sum_{k=1}^{L}d_k(\mu _n ^k - \mu _n)^2}$ denotes the inter-class variance and the denominator ${\sum_{k=1}^{L}d_k(\sigma _n ^k)^2}$ denotes the intra-class variance with respect to ${n_{th}}$ feature. Then the features are sorted in the order of higher Fisher score as features with higher Fisher scores tend to exhibit better differentiating capacity among classes. The Fisher score of each feature are normalized w.r.t. the maximum score and the logarithm of normalized Fisher score is plotted in \text{Figure~\ref{fig:correlation}}. It shows that the Fisher scores are higher for features having higher correlations. The best five features chosen from these two methods are summarized in Table \ref{tab:selectedfeatures}.

\subsection{\textbf{Feature Normalization:}}

\subsubsection{\textbf{Preparation of training data}}

The dataset is challenging to work with as failure causing features vary a lot and some  of  the  failure  states  have  similar  feature sets  as  those of active states,  making  it  infeasible  for neural networks to learn from these features directly. A commonly observed trend noted among the five selected features is that they have an increasing trend while approaching failure. So, min-max normalization i.e., normalization of features from 0 to 1 is applied to capture the increasing trend of features approaching failure on each data matrix ${A_i}$ having information about past 150 days.

A sequence length denoting the number of time steps to look back in the LSTM network (discussed later in this section) has been chosen with a trade-off between minimum required sequence length, time for training and online simulation and accuracy of prediction, and the data is organized as a three dimensional matrix 
(${Samples\times time steps\times features}$). The sequence length is varied and the one producing the minimum error between actual and predicted RUL is selected which in this case is twenty five days.

Each (${time steps\times features}$) matrix for each training example has the label of remaining useful life such that the network learns to predict RUL at any timestep given past 25 days of data inputs. The network is trained in a way to predict the RUL of a device within a range of 0 to 125 days from ${T_c}$, i.e., the current day when we decide to test a device. The training examples along with the labels are generated from which 5\% of the data are used for validation purposes. The training data has 71072 examples with the matrix of dimension ${(71072\times 25\times 5)}$.


\subsubsection{\textbf{Preparation of online simulation data}}

 For the training purposes, the data has been normalized from 0 to 1 with respect to ${T_f}$, the day when the device failed, while in case of simulation the actual feature statistics corresponding to failure is not known apriori. But as the increasing trend of features approaching failure has been observed, if a time series data for an active device with unknown RUL is taken then the maximum attainable value of a feature can be approximated from the historical data distribution. Considering that value to be maximum, the features in the given data matrix ${B_i}$ can be normalized and fed into the LSTM network to predict when it is going to fail. 


\begin{figure*}[h]
\begin{center}
    \makebox[\linewidth]{
\includegraphics[width=\textwidth]{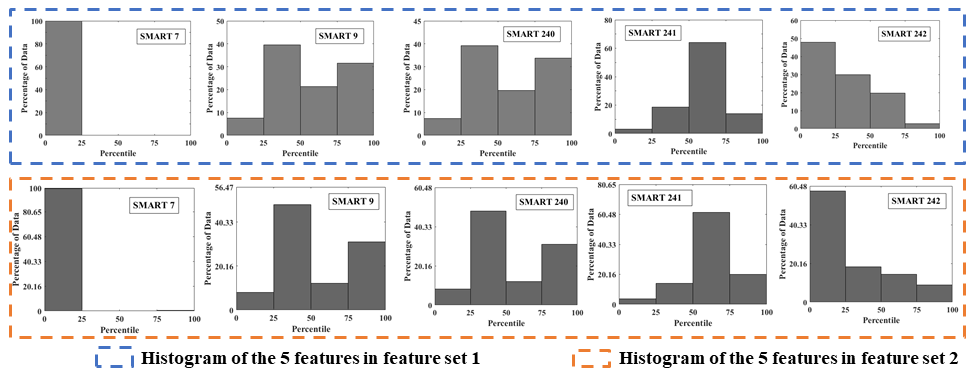}
}
\caption{This figure shows the histogram of five selected features (table \ref{tab:selectedfeatures}) and also emphasizes that the distribution of each feature for a duration of any two consecutive months are mostly similar. }
\label{fig:histogram}
\end{center}
\end{figure*}


In this process a lot depends on the values of features which will be taken as maximum and w.r.t. which the current feature values will be normalized.
For finding the optimal value for setting as maximum in the normalization process we look past two months' historical data distribution for any currently active device under test. In spite of fluctuations in the feature data within a shorter period of time, the overall moving average of feature values taking any data segment comprising of few months into account, tend to increase over time. Also the distribution of each feature for a duration of any two consecutive months are mostly similar as shown in \text{Figure~\ref{fig:histogram}}  where the histogram for historical data distribution of each feature looking back for two months from two different time instances referenced as feature sets 1 and 2 have been shown. So consideration of past two months of data in fetching historical maximum is appropriate in the sense that it is not too short to be affected by noisy fluctuations of small sequence of data as well as not too large so as to balance the computational overhead. Also, as the feature value variations have an underlying uptrend over months, so if too many months were considered in this scheme, then the prediction would have been biased towards values from those times of the year when the SMART values were even lower than the current values. This would have been an inaccurate representation and throw the predictions off by a margin.

\text{Figure~\ref{fig:histogram}} shows the histogram for each feature to show a typical distribution of that feature for any two consecutive months. On an average the histograms describe that feature 1 has most of its data concentrated within lower 25 percentile of the entire data range, with outliers occupying the rest of the dataspace. Features 2 and 3 have almost 70\% data within the upper quartile of the entire data spectrum. Feature 4 also has many outliers with 80\% of the total data within the 3rd quartile of the data spectrum. Feature 5 has 90\% of its data within the 3rd quartile of the data spectrum indicating a somewhat more sparse distribution in the 4th quartile. 

If the maximum value of each feature from historical data is considered as the value at failure for the simulation data, and feature values are subsequently normalized w.r.t. the historical maximum, the following happens: an approximation is made such that the feature values of the current (active) device under test will fail only when they reach the maximum attainable values given the past records. This is the most optimistic case. In this case, the normalized feature values will result in lower fractional values than it would have been, if normalized by its actual failing feature values, considering the maximum value of a feature may be an outlier in the distribution. This results into prediction of a much higher Remaining Useful Life (RUL) than in reality. This process of obtaining the best optimistic case RUL is referred to as Prediction Strategy 1.

\begin{figure}[h]
\includegraphics[width=\columnwidth,height=6cm]{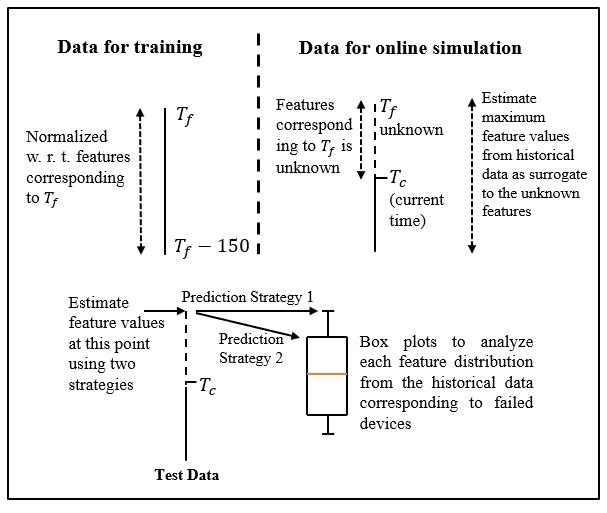}
\centering
\caption{Normalization of data for training and online simulation}
\label{fig:norm}
\end{figure}

To resolve this issue the data is sorted at first and normalized considering the values at the supremum of 3rd quartile (${75_{th}}$ percentile) of the entire data spectrum as the maximum feature value as most of the data is concentrated within this upper bound. Quartiles, as the measure of bounds, are chosen to work with to keep the generality of normalization over data from any part of the year. This is because in spite of having similar distributions, the data samples are not identically distributed as shown for feature sets 1 and 2 in \text{Figure~\ref{fig:histogram}}. Consequently, it is not possible to find a specific percentile that suits all. By choosing ${75_{th}}$ percentile, the upper bound stays close to the historical maximum yet the  elimination of significant number of samples of relevance to the analysis, can be avoided. This strategy also helps get rid of potential outliers in the sample sets. This approach to normalization, referred to as Prediction Strategy 2, results in significantly better approximations of the RUL. The normalization process is described in \text{Figure~\ref{fig:norm}} and Algorithm 1 describes the overall procedure for preprocessing and online prediction of RUL for the simulation stage. All symbols used are explained in \text{Table~\ref{tab:symbolsummary}}.

\begin{algorithm}
 \caption{Algorithm for preprocessing and online prediction of RUL for the simulation data}
 \begin{algorithmic}[1]
 \renewcommand{\algorithmicrequire}{\textbf{Input:}}
 \renewcommand{\algorithmicensure}{\textbf{Output:}}
 \REQUIRE 
 
 Simulation Data : \{${B_i\in \mathbb{R}^{m\times f}}$\} for each ${i}$
 \ENSURE  Predicted RUL for each ${i}$ from time ${T_c}$

  \STATE Load historical data ${H_{sh\times f}}$

  \FOR {$p = 1$ to $f$}
  
   \STATE{$\phi = sort (H(:,p))$}
   
   \STATE{${\phi_{75}} = sup({Q_3}(\phi))$}
   
   \STATE{${\Tilde{B_i}(:,p)}$ = (${B_i}$(:,p) - min(${B_i}$(:,p))) / (${\phi_{75}}$ - min(${B_i}$(:,p)))}
  
 \ENDFOR

 \STATE ${L = \Tilde{B_i}(1:ts,:)}$

 \STATE Feed L to the pretrained LSTM network as online simulation data
 \STATE Output the RUL 
 
 \end{algorithmic}
 \end{algorithm}

\begin{figure}[ht!]
\includegraphics[width=\columnwidth,height=5cm]{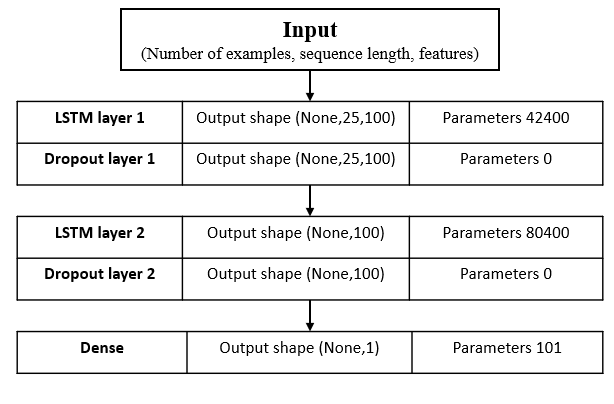}
\centering
\caption{The Stacked LSTM Architecture}
\label{fig:stackedlstm}
\end{figure}


\section{\textbf{Long Short Term Memory (LSTM)}}

\label{sec:lstms}

LSTM is a widely used variant of recurrent neural network proposed by Hochreiter and Schmidhuber \cite{doi:10.1162/neco.1997.9.8.1735}, capable of capturing dynamic temporal behavior in time series data by the use of shared parameters while traversing through time. A self-feedback LSTM unit associated with input, output and forget gate, control the motion of information through the gating mechanisms. As the name suggests, the input and output gate of each memory cell in LSTM directs the inputs and outputs flowing in and out from the cell respectively, whereas the forget gate decides upon which information needs not to be memorized anymore. The values of the gates are decided by sigmoid activation function. At any point of time it takes the current input, captures hidden state information for previous time steps upto a given sequence length and generates output according to the task given.
The input feature at a timestamp combining current input, previous hidden state information and shared parameters with ${tanh}$ activation function is multiplied with the input gate coefficient and updates the value of the memory cell combining the forget gate coefficient controlled previous timestamp\textquotesingle s value of the cell. Any hidden output state is thus a result of output gate-controlled ${tanh}$ activated memory cell value, which is then used for producing outputs.
 \text{Figure~\ref{fig:stackedlstm}} shows the stacked LSTM architecture.



\begin{figure}[ht!]
\includegraphics[width=\columnwidth,height=5cm]{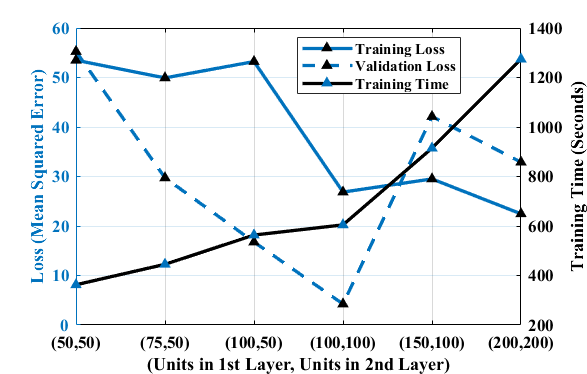}
\centering
\caption{Selection of optimal number of units per layer based on the training loss, validation loss and training time}
\label{fig:unitvstime}
\end{figure}

\begin{figure}[h]
\includegraphics[width=\columnwidth,height=5cm]{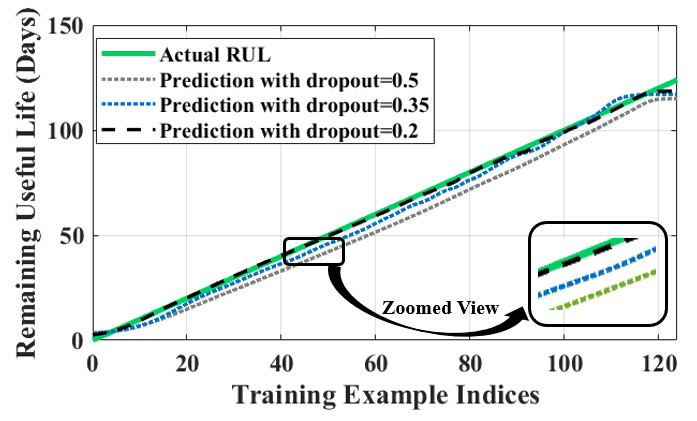}
\centering
\caption{Effect of variation of dropout ratio on estimation of RUL}
\label{fig:dropoutratio}
\end{figure}

\subsection{\textbf{Hyper-parameter Selection}}

\begin{itemize}[leftmargin=*]
    \item \textbf{Number of Units in each layer:} For selecting optimal number of units for two layered LSTM model a graph showing varying number of units per layer and training and validation loss corresponding to them along with execution time is plotted as presented in \text{Figure~\ref{fig:unitvstime}}. From the figure it is evident that the optimal number of units per layer is 100 with quite low execution time as increasing number of units further increases validation loss much greater than training loss indicating overfitting.
    \item \textbf{Selection of Dropout Ratio:} 
Dropout is an important regularization parameter to control over-fitting. 
The effect of variation of dropout ratio on the accuracy of estimation of Remaining Useful Life has been shown in \text{Figure~\ref{fig:dropoutratio}}. According to the results we chose dropout ratio of 0.2 for both the layers.
\end{itemize}



To prove the fit of LSTM to this problem we also trained a Naive Bayes classifier which is a probabilistic classifier based on Baye's Theorem. The network has good training and validation accuracy because of the normalized structure of training data and distinct labels associated with distinct feature values. But at the time of testing for devices it fails completely.
This is due to the fact that the combination of features for a given input of simulation data do not necessarily match with the combinations with which it was trained. Therein, comes the utility of of LSTM, where the model is being able to predict the labels learning the temporal embeddings of failure. 


The online prediction model used in our approach is computationally inexpensive as we are just feeding the simulation data into pre-trained LSTM model and obtaining the remaining useful life instantly. The network takes approximately 0.005 seconds to predict the RUL of a device on insertion of processed simulation data into the pre-trained LSTM unit.

\section{Results and Discussion}

\label{sec:results}

\subsection{\textbf{Beyond training and cross-testing}}

\begin{figure}[h]
\includegraphics[width=9.7cm, height=3.3cm]{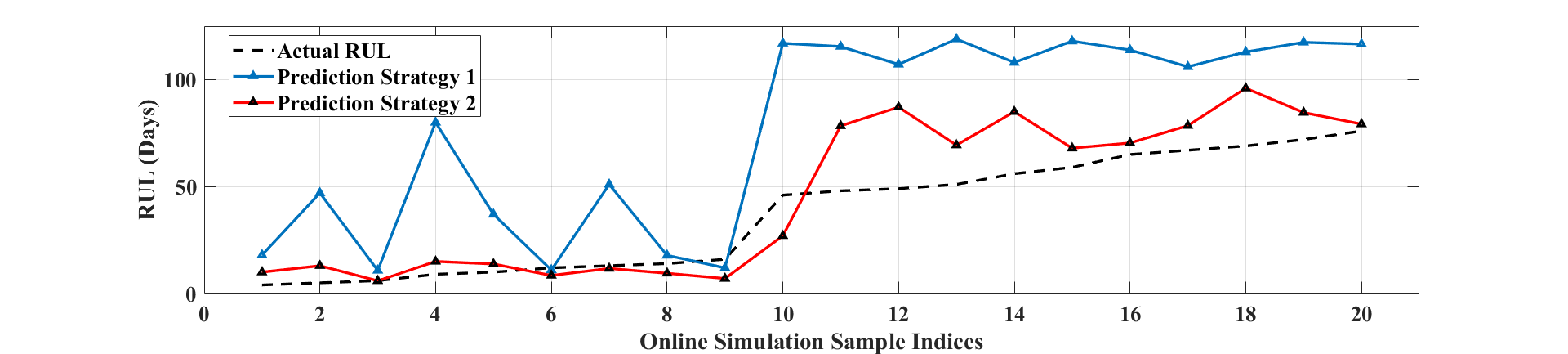}
\centering
\caption{The actual and predicted Remaining Useful Lives of various devices under online simulation}
\label{fig:predstrategy12}
\end{figure}

The preparation of online simulation data is described in Algorithm 1 and \text{Figure~\ref{fig:norm}}. 
The actual and predicted RUL of the devices are shown in \text{Figure~\ref{fig:predstrategy12}} where different disk drives having various remaining useful lives have been considered to show the performance of the proposed approach over a diverse range of simulation example indices. 

Overall, prediction strategy 2 produces better estimation of RUL than prediction strategy 1 for the devices with impending failures which is more important and provide a fair approximation of devices that are going to fail later.


\begin{figure}[h]
\includegraphics[width=9cm,height=3.7cm]{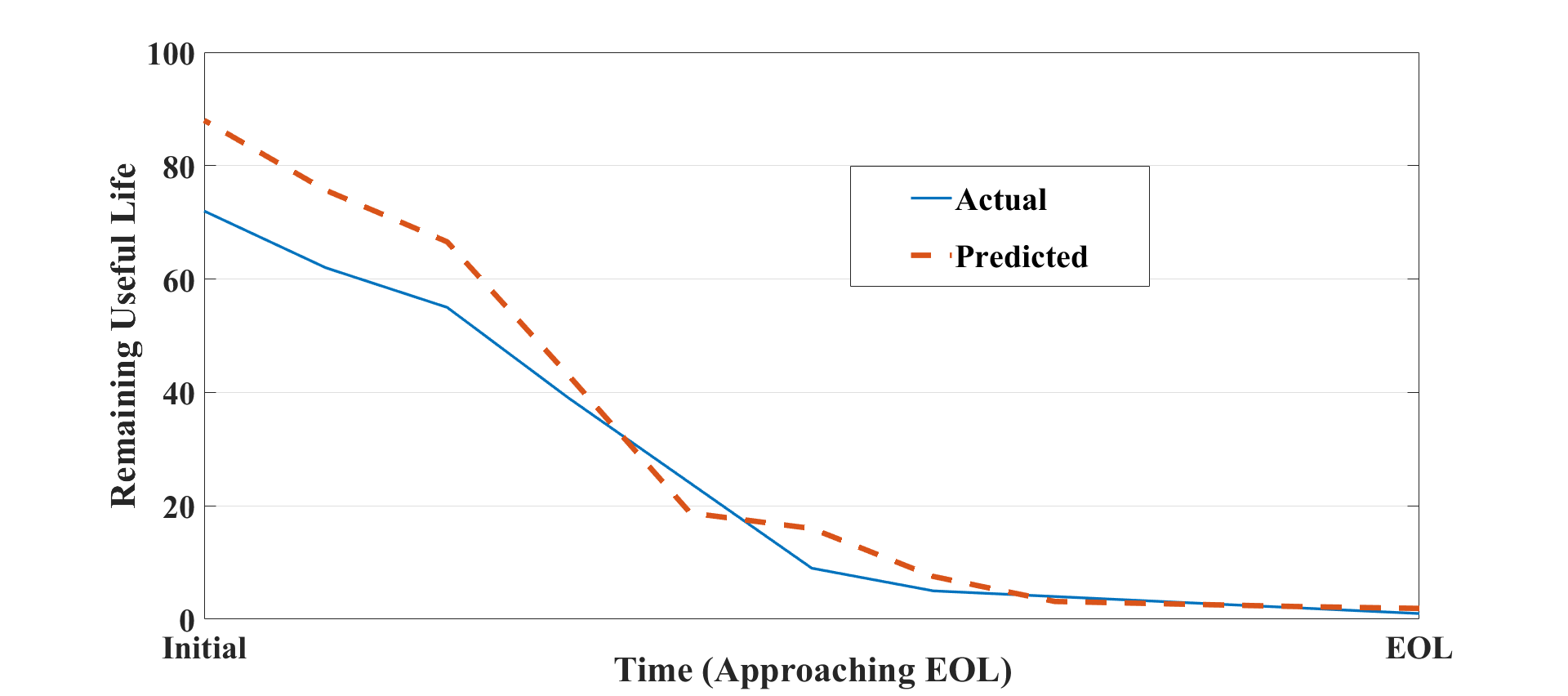}
\centering
\caption{Reduction in Uncertainty in Prediction As a Device Approaches EOL}
\label{fig:uncertainty}
\end{figure}



\begin{figure}[h]
\includegraphics[width=9cm, height=3.8cm]{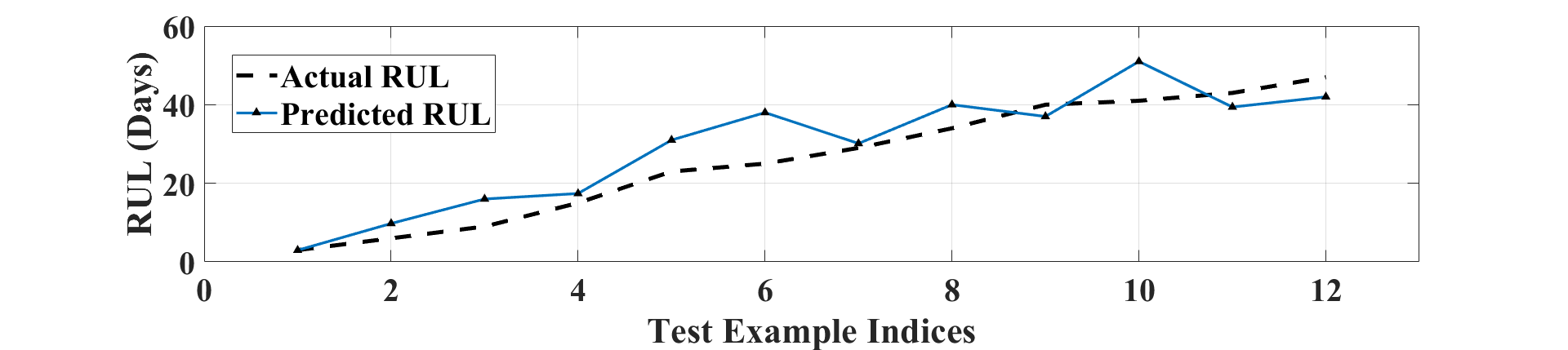}
\centering
\caption{The actual and predicted Remaining Useful Lives of various devices under online simulation for a different model from same manufacturer}
\label{fig:othermodel}
\end{figure}

\subsection{\textbf{Variation of Predicted RUL over time for a single disk}}

Generally in a cloud service system, the revenue loss caused due to allocating Virtual Machines to faulty hard disks can be modeled as a sum of the false positives and negatives weighted by their respective losses.
\begin{equation}
Revenue Loss = loss_{nfp}\times nfp + loss_{nfn}\times nfn
\end{equation}

Where ${loss_{nfp}}$ and ${loss_{nfn}}$ are losses incurred due to false positives ${nfp}$ and false negatives ${nfn}$ respectively. Depending on the application, more importance can be given to precision or recall based on the costs associated with ${loss_{nfp}}$ and ${loss_{nfn}}$. 

As we are more interested in predicting imminent failures, the prediction accuracy for devices which are going to fail sooner is more critical than those that having greater RUL. Based on the online prediction results of disk drives having imminent failure the decision on the allocation of jobs in the cloud architecture is to be taken. So the prediction accuracy is much sensitive in this region as smaller number of false positives or false negatives can incur greater revenue loss. It is of critical importance to determine how the uncertainty in prediction of RUL for any device reduces as it approaches its end of life and remains to be analyzed. \text{Figure~\ref{fig:uncertainty}} shows a graph of prediction of RUL for a single device at different time instances and indicates greater accuracy of prediction near the time region of actual failure.

\subsection{\textbf{Variation of Precision, Recall and F1 score over time}}

\begin{figure}[h]
\includegraphics[width=\columnwidth]{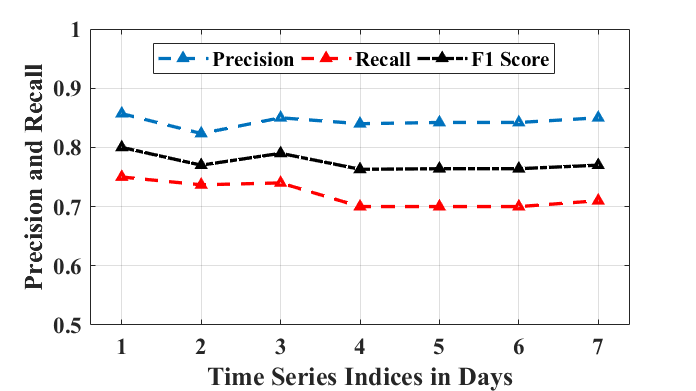}
\centering
\caption{Precision and Recall of Prediction of RUL over Seven Consecutive Days}
\label{fig:precisionandrecall}
\end{figure}

The Precision and Recall of prediction of RUL for numerous devices have been shown in \text{Figure~\ref{fig:precisionandrecall}} based on the fixed threshold of whether a device is going to fail in next ten days. The process is carried on for seven consecutive days to get a time series variation of these measures over a fixed threshold. The plot shows an average Precision of 0.84, Recall of 0.72 and F1 score of 0.77.
The flat nature of the curve indicates  the consistency  and  robustness  in  decision  making using the model over  several  consecutive  days.

\subsection{\textbf{Generalizable and transferable architecture among different disk models}}

One of the major advantages of the proposed online RUL prediction system is the ability to transfer the architecture learned from one disk model to others. The network was trained with model ST4000DM000 and the pre-trained model on ST4000DM000 has been tested on a different model ST8000DM002 from Seagate which provided acceptable results on prediction of RUL over simulation example indices as shown in \text{Figure~\ref{fig:othermodel}} using prediction strategy 2. Hence the proposed framework can be trained on a single model from a manufacturer and can be successfully used to predict RUL for different models from same manufacturer proving its generalizability and transferability. The only thing to keep in mind is that if the historical data distribution of a feature varies drastically from the model the network has been trained on, a different threshold for the normalization step at the time of online simulation might produce more accurate result.

\section{Comparative Analyses with existing research}

\label{sec:comparative}

Despite the significant existing work on the issue, there is a lack of an overall framework for online device health monitoring that leverages the learning capabilities of LSTM Networks extracting meaningful information from the sequences of data to identify trends indicating a device approaching failure. Further, through our generalized prediction framework we show that our proposed architecture trained on one model from a manufacturer performs well on the other models from the same manufacturer, which has not been shown explicitly in the existing research in this domain.

\begin{figure}[h]
\includegraphics[width=\columnwidth, height=4.4cm]{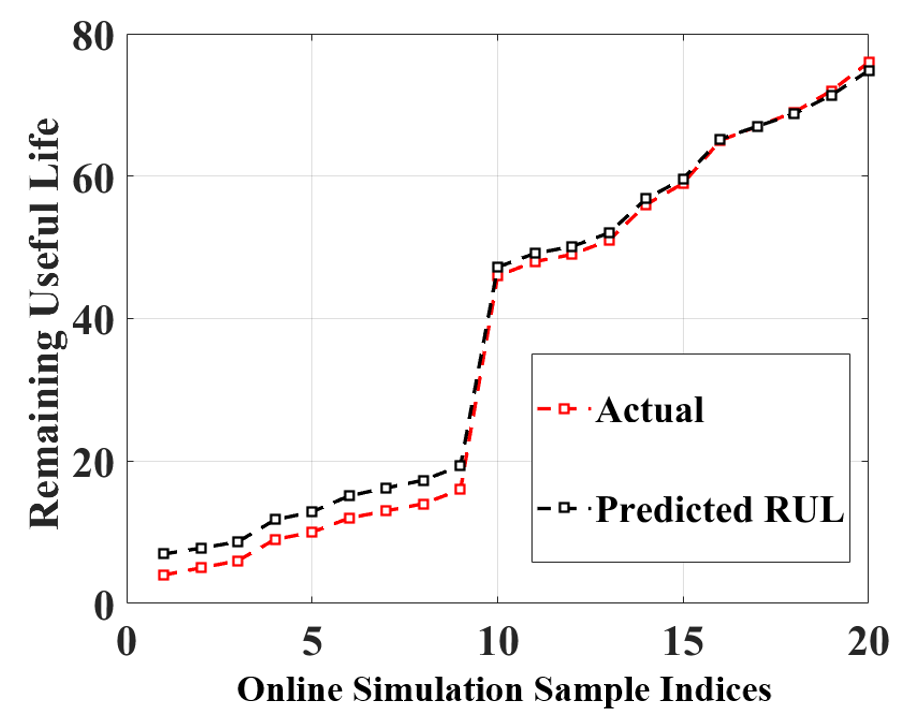}
\centering
\caption{Evaluation Results using future Information in online simulation}
\label{fig:testing}
\end{figure}

In most of the existing research \cite{8457760} \cite{203217}  the entire dataset is first normalized and then divided into training and test sets which are drawn from same distribution. In this way, the information of failure gets embedded in test data. This does not represent real world test cases if we want the model to perform online prediction with simulation data without any knowledge on failure characteristics. If we would have drawn the simulation data distribution in a similar fashion as that of the training one, using future information, then we could have had much better prediction of the RUL as shown in \text{Figure~\ref{fig:testing}}. But as this does not indicate the actual efficiency of the prediction model, this cannot be used in a real life scenario. In \cite{8260700} the authors claimed that the best performance on Backblaze dataset was shown by Random Forest (RF). The Precision and Recall values based on the threshold of device failure within 10 days were recorded as almost 0.93 and 0.6 dividing the dataset into training and test sets using cross-validation techniques, whereas we obtained an average precision of 0.84 and recall of 0.72 using the decision threshold of device failure within 10 days without using any future information in the simulation process. Hence our proposed architecture is able to mitigate the challenge of predicting RUL of a device without any future knowledge of online simulation data with acceptable decision outputs manifesting real-world scenarios.

In \cite{oldlstm} deep neural network based RUL estimation has been carried out by both LSTM and CNN where LSTM achieved better performance. The authors chose to perform the tests on the same Seagate model ST4000DM000 and performed the prediction of RUL on models of a single serial number Z300ZQST where the RMSE and the deviation of predicted result from ground truth are shown. In our work the trained network on model ST4000DM000 showed satisfactory performance on a completely different model from Seagate covering various serial numbers. We also discuss separate preprocessing steps for training and online simulation data so as to emulate the real scenario.

\section{Conclusion and future work}

\label{sec:conclusions}

In this work, we proposed a data-driven framework using deep LSTM architectures for online estimation of remaining useful life of devices where the feature values corresponding to  failure  are  not  uniform across devices.  The architecture proposed is efficient in predicting imminent device failures and is generalizable and transferable to different disk models. Although, the inferences made are based on our work on the hard disk data, the combined Normalization and Inference Mechanisms shown in this paper are applicable to any generic timeseries progressing towards failure. In future, we will extend this architecture to create an adaptive job scheduler which considers hardware failures, workload forecasts and application interface models. We expect to use this generalized framework in various related applications with data specific modifications focused on online relaying of decisions of critical importance.

\bibliographystyle{IEEEtran}
\bibliography{IEEEabrv,Bibliography}

\end{document}